# Deep neural networks can be improved using human-derived contextual expectations

Harish Katti, Marius V. Peelen, and S. P. Arun

- *Harish Katti is with the Centre for Neuroscience, Indian Institute of Science, Bangalore, India, 560012. E-mail: [harish2006@gmail.com](harish2006@gmail.com) (to whom correspondence should be addressed)*

- *Marius V. Peelen is with the Donders Institute for Brain, Cognition and Behaviour in Nijmegen, The Netherlands. E-mail: m.peelen@donders.ru.nl.*

- *S. P. Arun is with the Centre for Neuroscience, Indian Institute of Science, Bangalore, India, 560012. E-mail: sparun@iisc.ac.in*

| | |
|---|---|
| Abbreviated title | : Human contextual expectations improve deep networks |
| Number of Figures | : 5 |
| Number of Tables | : 3 |

# ABSTRACT

Real-world objects occur in specific contexts. Such context has been shown to facilitate detection by constraining the locations to search. But can context directly benefit object detection? To do so, context needs to be learned independently from target features. This is impossible in traditional object detection where classifiers are trained on images containing both target features and surrounding context. In contrast, humans can learn context and target features separately, such as when we see highways without cars. Here we show for the first time that human-derived scene expectations can be used to improve object detection performance in machines. To measure contextual expectations, we asked human subjects to indicate the scale, location and likelihood at which cars or people might occur in scenes without these objects. Humans showed highly systematic expectations that we could accurately predict using scene features. This allowed us to predict human expectations on novel scenes without requiring manual annotation. On augmenting deep neural networks with predicted human expectations, we obtained substantial gains in accuracy for detecting cars and people (1-3%) as well as on detecting associated objects (3-20%). In contrast, augmenting deep networks with other conventional features yielded far smaller gains. This improvement was due to relatively poor matches at highly likely locations being correctly labelled as target and conversely strong matches at unlikely locations being correctly rejected as false alarms. Taken together, our results show that augmenting deep neural networks with human-derived context features improves their performance, suggesting that humans learn scene context separately unlike deep networks.

## INTRODUCTION

*We work with being, but non-being is what we use.*

*- Tao Te Ching* [1]

Detecting targets in real world scenes remains a hard problem even for the hugely successful deep convolutional neural networks (CNNs). For instance, state-of-the-art deep convolutional networks can detect people with 82-88% accuracy and cars with 77-84% top-1 accuracy [2,3] based on our evaluation on a real world scene dataset [4], whereas humans fare much better on comparable scenes at 93% [5]. One potential reason for this performance gap is that humans and machines have qualitatively different training data. Machines are typically trained on large image databases containing targets embedded in their surrounding context. This can compromise their ability to learn useful context features in the presence of vastly more useful target features. In contrast, humans often see scenes in which the target object moves out of view or moves against a static background. This provides an opportunity for humans to learn separate features for target and context. If this is true, it follows that humans must have systematic expectations about target objects even on scenes that do not contain those targets. The absence of targets provides an opportunity for machine algorithms to learn the weaker context signals that drive human expectations. If state-of-the-art object detectors have indeed learned only target-related signals, it follows that their performance can be improved by augmenting them with human-derived contextual expectations.

That context can aid in object detection has been appreciated both in studies of human vision as well as computer vision. In humans, it is well known that finding objects in a congruent context is faster than in incongruent contexts [6,7]. Brief previews of scenes have been shown to guide eye movements towards cued targets [8]. Both nontarget objects and coarse scene layout contribute to object detection [9–12] although their relative contributions have only been elucidated recently [5]. In the brain, there are dedicated scene processing regions [13] that respond to scenes as well as to their associated objects [14,15]. In computer vision, contextual priors learnt from target present scenes have been used to improve object detection and localisation by constraining the locations to search [16–18]. Models incorporating contextual features have also been shown to be useful in predicting task directed eye-movements [19]. More recently, deep convolutional networks have shown dramatic improvements in scene [20] and object classification [21]. However, it is not clear whether these deep networks learn target and/or context features. Thus, while there is evidence that scene context can facilitate object detection in both machines and humans, it is largely thought to facilitate searching for objects. Furthermore, whether context involves processing target features, associated nontarget objects, and/or scene layout has remained unclear.

Here we set out to investigate whether humans form systematic expectations on scenes that do not contain target objects, and asked whether these expectations can be understood and predicted using computational modelling. By learning these expectations, we were able to generate predicted human expectations on much larger datasets without requiring laborious manual annotation. We then asked whether augmenting state-of-the-art object detectors with these human-derived expectations improves overall performance.

# RESULTS

Our central premise was that humans have access to separate target and context feature representations. We selected cars and people as suitable test categories because they are ecologically important, extensively researched [22–24] and common in popular datasets [25–28]. Our results are organized as follows: We first performed a behavioural experiment on humans in which we measured their contextual expectations on natural scenes and used computational modelling to understand and predict these expectations. Second, we demonstrate that these predicted human expectations can be used to significantly improve the performance of state-of-the-art object detectors. Finally, we demonstrate that this improvement is non-trivial in that it cannot be obtained using target-related signals of various types. To facilitate further research, the code, behavioural data, visual features and stimuli used for this study are publicly available at https://github.com/harish2006/cntxt_likelihood.

**Measuring human expectations (Experiment 1)**

If humans can process object features independently of context, then they must be able to form systematic expectations about the likelihood, scale and location of where objects might occur in a scene. Here we set out to measure these expectations systematically using a behavioural experiment on human subjects. On each trial, subjects were shown a scene that did not contain cars or people, and were asked to indicate the likelihood, scale and location of where cars or people might occur in the scene at a later point in time (see Methods for details).

Figure 1 illustrates the systematic expectations produced by humans on two example scenes: the first scene was rated by human subjects as likely to contain people but not cars, whereas the second was rated as likely to contain cars but not people. To measure the reliability of these expectations, we divided the subjects into two groups and calculated the correlation between the average rating obtained from each group. All correlations were large and highly significant (r = 0.94, 0.9, 0.91, 0.89, 0.47 for likelihood, x-position, y-position, area and aspect ratio respectively between odd- and even-numbered subjects for cars; r = 0.87, 0.79, 0.96, 0.86 & 0.36 for people; p < 0.00005 for all correlationss).

**Computational models for car and person likelihood**

Next we asked whether the above systematic expectations can be predicted and understood using computational modelling. To this end we divided the image features present in each scene into target-related features, non-target objects and scene context features (see Methods). The inclusion of target-related features might appear counter-intuitive at first glance since these scenes do not contain target objects. However we included them nonetheless for completeness as well as because human expectations might still be driven by the weak presence of target-like features in a given scene. We tested a number of models based on combinations of target, nontarget and coarse scene information. Models were evaluated for their ability to predict the average likelihood ratings for novel scenes that were never used in model fitting (Table 1).

Overall, the best model for likelihood ratings was the one containing nontarget and coarse scene but not target features. We determined it to be the best model because (1) it yielded better fits to the data than models trained with only target, nontarget or coarse scene features (p < 0.001 in all cases). (2) It outperformed models based on other pairs of feature channels i.e. target and nontarget (p < 0.001 in both cases) or target and coarse scene structure (p < 0.01 in both cases) (3) its performance was equivalent to the full model containing target, nontarget and coarse scene features (p > 0.05). All values are given in Table 1. The performance of the best model is illustrated along with example scenes in Figure 2.

We then asked whether nontarget objects which increase car likelihood, also decreased person likelihood and vice-versa. For this analysis, we extracted regression weights for nontarget object labels in models that predicted person likelihoods and plotted them against regression weights for the same nontarget labels in models that predicted car likelihood. We obtained a negative and significant correlation confirming this prediction ($r = -0.31$, $p < 0.05$). We observed that nontargets such as *signage*, *cables* that frequently occur on highways tend to increase car likelihood and decrease person likelihood. Conversely, nontarget labels such as *bench*, *stair* and *cycle* tend to increase person likelihood and decrease car likelihood. Both patterns are as expected given the associations of these objects with cars and people respectively.

| Model Name | Correlation with person likelihood | Correlation with car likelihood |
|---|---|---|
| *Ceil* | *0.87±0.02* | *0.94±0.01* |
| TNC | 0.65±0.01# | 0.59±0.01# |
| T | 0.21±0.02* | 0.12±0.02* |
| N | 0.51±0.02* | 0.53±0.01* |
| C | 0.61±0.01* | 0.48±0.01* |
| TN | 0.54±0.02* | 0.52±0.01* |
| TC | 0.60±0.01* | 0.47±0.01* |
| NC | **0.65±0.01** | **0.60±0.01** |

**Table 1. Model performance on predicting car/person likelihood ratings in humans.** *Ceil* refers to data reliability, which is an upper bound on model performance given the inter-subject variability in ratings (see text). The best model for predicting car and person likelihoods was based on nontarget and coarse scene features (NC). We calculated model performance as the average cross-validated correlation (*mean ± sd*) over 1000 random 80-20 splits of the scenes. Asterisks represent the statistical significance of the comparison with the NC model (* is $p < 0.001$, # is $p > 0.05$). Statistical significance was calculated as the fraction of 1000 random 80-20 splits in which model correlation exceeded the best model. Note that model performance sometimes reduces after adding extra features because of overfitting. Abbreviations: T, N, C: Targets, Nontargets and Coarse features. TN = Targets & Nontargets, etc.

**Computational modelling of likely location, scale and aspect ratio**

Next we asked if models based on combinations of target, nontargets and coarse scene features could predict other aspects of the likelihood data, namely the average horizontal location, vertical location, scale (i.e. area) and aspect ratio (i.e. vertical/horizontal extent) indicated during the likelihood task by human subjects. We visually inspected the annotated boxes that that subjects had drawn to indicate likely car or person locations and found that the average horizontal or vertical locations are meaningful in all but few exceptions such as when subjects draw boxes corresponding to likely person locations on either of two deck chairs and the average person box ends up being in the middle of two chairs. The results are summarized in Supplementary Table 1. In general models containing nontarget and coarse scene information (NC) yielded the best predictions (Figure 2). Model predictions were significantly correlated with the observed human data but, fell short of the noise ceiling (Figure 2), indicating differences in the underlying features used by humans and models.

Interestingly, models were better at predicting the vertical position of cars or people compared to horizontal location. This could be because vertical locations of cars/people vary less than horizontal locations, or because horizontal locations are harder to predict since its variations are due to differences in 3d scene layout. We note that the difficulty of predicting horizontal object locations has been reported previously [16].

**Comparison with other computer vision models**

To confirm the validity of our models and the specific choice of the feature channels, we compared the performance of the best model (NC) with the performance of three other models: (1) a pixel-based model in which image pixels are used directly as input; (2) a CNN pre-trained for 1000-way object classification [29] and (3) a CNN pre-trained for scene classification [30]. The NC model yielded qualitatively similar but slightly lower performance compared to the CNNs on predicting likelihoods, vertical position and scale but was better able to predict the expected horizontal location of targets (Figure 3). In this manner we find that the NC model predicts human expectations well and offers benefits of model simplicity and interpretability. All model predictions again fell short of the noise ceiling of the human data, indicating systematic differences in the underlying feature representations between models and humans.

**Augmenting deep networks with human-derived context expectations**

The above results show that humans form highly systematic expectations about the overall likelihood, location and scale at which cars or people might occur in a scene, and that these expectations are largely driven by coarse scene features and the presence of nontarget objects. The fact that human expectations could be predicted using computational modelling meant that we could use these models to generate predicted human expectations without requiring any laborious manual annotations from human subjects.

In this section, we evaluate the central premise of our study, namely that if deep networks have learned the stronger target features at the expense of the weaker context signals, augmenting them with human-derived context features should lead to substantial gains in performance.

We trained linear classifiers using feature vectors formed by concatenating confidence score from each CNN for the target category together with the predicted human expectations (likelihood, horizontal and vertical positions, scale and aspect ratio) generated for novel scenes without human annotations. To generate these predictions, we used the context-only model for multiple reasons. Firstly, because it explains most of the variance in the human ratings (Tables 1, Supplementary Table 1). Secondly, since inputs to the context-only model are highly blurred scenes, any improvement arising from predicted human priors can be attributed purely to contextual guidance and not target/object related information. Thirdly, it is trivial to extract coarse scene features from novel scenes but is impractical to assume the availability of manually annotated nontarget labels for the same. The resulting model performance is summarized in Table 2. It can be seen that the augmented models perform uniformly better with better performance on scene categories shared with our original dataset. The improved accuracy was not merely a result of adding more parameters since the accuracy is cross-validated (Table 2).

| CNN | Target | CNN | CNN + Lklhd | CNN + yLocn | CNN + scale | CNN+ Lklhd + yLocn + scale | CNN + all car & person ratings | Increase in % |
|---|---|---|---|---|---|---|---|---|
| RCNN | C | 88.3 | 88.3 | 88.3 | 88.3 | 88.4 | **88.6** | **0.3** |
| RCNN | P | 84.0 | 84.6 | 84.0 | 84.0 | 84.8 | **84.9** | **0.9** |
| RCNN | CM | 82.4 | 85 | 82.5 | 83.2 | 85.5 | **86.2** | **3.8** |
| RCNN | PM | 80.6 | 80.6 | 81.5 | 80.6 | 80.4 | **82.0** | **1.4** |
| Alexnet | C | 82.5 | 82.6 | 82.4 | 83.0 | 83.3 | **84.0** | **1.5** |
| Alexnet | P | 76.8 | 77.8 | 76.7 | 76.8 | 78.2 | **78.7** | **1.9** |
| Alexnet | CM | 83.5 | 85.8 | 83.5 | 84.3 | 86.8 | **87.1** | **3.6** |
| Alexnet | PM | 73.4 | 73 | 77.1 | 75.0 | 76.8 | **77.1** | **3.7** |

**Table 2: Improvement in car/person detection obtained by augmenting state-of-the-art CNNs with predicted human-derived contextual expectations.** Each entry shows the cross-validated accuracy for detecting cars on all (C) or matched (CM) scenes from ADE20K, or people on all (P) or matched (PM) scenes from ADE20K – for details see Supplementary Table 2. The matched scenes comprised scene categories similar to those rated by humans. Best models are highlighted in bold. Columns indicate the kind of model used: the column marked CNN indicates the baseline accuracy of the deep neural network; the columns of the form "CNN + X" indicate accuracy for CNN augmented with feature X. Lklhd: predicted likelihood of target category object; xLocn: predicted horizontal location of target category object, yLocn: predicted vertical location of target category object; scale: overall bounding box area marked by subjects.

Example scenes that contain cars at scales and locations that make them hard to detect reliably are shown in Figure 4a. These scenes were classified correctly by augmenting CNN decisions with human derived priors. We find that scenes with box like objects can result in false alarms for cars, that are then effectively suppressed by incongruent scene layouts such as the abbey tower, building façade and bar counter scenes (Figure 4b). Likewise, we find CNNs miss out people in many scenes (Figure 4c) when people are present at very small scales or eccentric locations, such scenes also benefit from augmentation. Like in the case of cars, we find that incongruent contexts can also suppress false alarms like in the case of the shopping mall scene or outdoor farm scene with a tractor (Figure 4d), in both cases the presence of people at large scales is ruled out.

To further elucidate why CNN accuracy is benefited by augmenting with human contextual expectations, we first chose the restricted set of 372 car scenes (Table 2, third row) and plotted the predicted car likelihood for each scene against the car CNN confidence scores obtained from [2] (Figure 5a). The augmented classifier boundary has a negative slope that results in better performance. This performance improvement can be attributed to weak matches on high-likelihood scenes being correctly declared as targets, and strong matches on low-likelihood scenes being correctly rejected as a non-target. This improvement can be seen also in the ROC curves obtained by varying the decision criterion for the original CNN and the augmented CNN (Figure 5b). We obtained results on augmenting CNN person scores from [2] for the restricted set of 306 person scenes (Table 2, fourth row) with predicted vertical location. We obtained qualitatively similar benefits as in the case of cars (Figure 5c-d; Table 2).

We found that both RCNN and Alexnet were conservative and rejected disproportionately more target present scenes than incorrectly classify target absent scenes. For RCNN, we observed 80 incorrect person rejections compared to 38 person false alarms on the matched person set *PM* and 126 incorrect car rejections compared to 8 car false alarms on the matched car set *CM*. For Alexnet, we observed 177 incorrect person rejections compared to 20 person false alarms on the matched person set *PM* and 124 incorrect car rejections compared to 10 car false alarms on the matched car set *CM*). So, greater benefits of augmentation can be expected due to boosting of weak target evidence in scene layouts that strongly suggest the presence of target objects.

**Does augmenting improve accuracy on other categories as well?**

Could augmenting CNNs with car/person expectations improve accuracy on other categories as well? This is plausible since many objects (e.g. bottle, train) are strongly associated with people. We tested this idea by augmenting CNN confidence scores for a number of additional categories with predicted car/person expectations as before. Remarkably, we obtained an improvement in classification accuracy of 3-20% on a number of categories from the Pascal VOC challenge set [31] (Table 3), on scenes that closely matched our reference set of 650 car-person absent scenes (scene categories detailed in [5]). Since many of these classes are rare even in the large sized ADE20K dataset[4], our results show that augmenting with human priors can provide benefits beyond the categories for which human annotation was obtained and amortize the effort needed to obtain human priors for few categories.

| Category | #Scenes | Alexnet | Alexnet + Car & Person ratings | Improvement in % | RCNN | RCNN + Car & Person ratings | Improvement in % pts |
|---|---|---|---|---|---|---|---|
| Airplane | 77 | 81.0 | 81.0 | **0.0** | 68.8 | 72.5 | **3.7** |
| bicycle | 261 | 64.5 | 77.5 | **13.0** | 54.4 | 73.4 | **19.7** |
| Bird | 40 | 52.3 | 71.8 | **19.4** | 58.8 | 65.7 | **6.9** |
| Bottle | 13 | 59.2 | 77.8 | **18.5** | 79.5 | 79.5 | **0.0** |
| Bus | 202 | 53.7 | 67.8 | **14.2** | 75.1 | 77.7 | **2.8** |
| Chair | 14 | 85.7 | 89.4 | **3.7** | 68.8 | 68.8 | **0.0** |
| Dog | 53 | 58.8 | 68.3 | **9.5** | 43.6 | 63.8 | **20.2** |
| Horse | 22 | 68.8 | 70.5 | **1.7** | 45.8 | 51.7 | **5.9** |
| Motor | 20 | 62.4 | 78.9 | **16.5** | 70.8 | 79.4 | **8.6** |
| Pot | 139 | 54.5 | 74.2 | **19.7** | 84.5 | 89.3 | **4.8** |
| Couch | 402 | 83.8 | 85.9 | **2.1** | 83.0 | 85.4 | **2.4** |
| Train | 21 | 75.8 | 77.5 | **1.7** | 76.7 | 76.7 | **0.0** |
| Tv | 226 | 73.4 | 80.1 | **6.7** | 80.0 | 83.7 | **3.7** |

**Table 3: Improvement in accuracy for other object categories.** Here too, two types of CNN object detectors: Alexnet [3] and RCNN [2] were augmented using human-derived car/person likelihood scores on novel scenes.

Why do some categories benefit by augmenting with human-derived expectations but not others? We discovered two systematic patterns. First, categories with low baseline CNN performance might benefit by augmentation. Indeed, there was a significant positive correlation between the augmentation benefit and baseline CNN accuracy (r = 0.53, p = 0.05 across the 13 PASCAL VOC [31] categories tested). Second, categories strongly associated with people or cars – such as bicycle – might benefit by augmenting with human-derived people/car expectations. To assess this possibility, we calculated for each category the conditional probability of it occurring when a car was also present: p(object present|car present). If that object is associated with the presence of a car, its probability will be larger or smaller than the probability p(object present) across the dataset. We took the absolute difference between these two quantities therefore as a measure of association between each category with cars, and likewise calculated a similar association index for people as well. The average association index (across cars and people) was significantly correlated with the augmentation benefit (r = 0.68, p < 0.005 across 13 categories). Thus, objects that are strongly associated with cars and people experience a greater benefit by augmenting with human expectations for cars and people.

**DISCUSSION**

We started with the premise that the standard approach of training machine vision algorithms on scenes with target objects in their natural scene context will lead to learning stronger signals from the target object at the expense of the weaker context signals. In contrast, we reasoned that humans have the opportunity to learn separate target and context signals due to their visual experience. Here we confirmed this premise by demonstrating that (1) Humans form systematic expectations about the likelihood, scale and position of potential target objects in scenes entirely lacking the object of interest; (2) These expectations can be learned using computational modelling, and can be used to augment state-of-the-art CNNs to improve performance; (3) This improvement was due to relatively poor matches at highly likely locations being correctly labelled as target and conversely strong matches at unlikely locations being correctly rejected as false alarms; and (4) This benefit is non-trivial in that it cannot be obtained by simply augmenting CNNs with other types of human responses or other computational models.

The fact that state-of-the-art object detectors can be improved by augmenting them with human likelihood ratings raises several interesting questions. First, what about augmenting object detectors directly with human performance during object detection itself? Human priors have been studied previously using gaze locations recorded while people search for targets. In these tasks [19], more fixations are observed when people take longer to find the target, and these fixations can be predicted using scene gist. This raises the possibility that learning from human behaviour (eye position/response times) during object detection could produce similar gains in performance as observed with the human likelihood ratings. To address this issue, we used data from a previous study in which we measured the response times of humans during target detection on the same scenes [5]. Interestingly, observed response times were uncorrelated with observed car likelihood ratings ($r = 0.005$, $p = 0.9$) and only weakly correlated for person likelihood ratings (r = 0.2, p < 0.005). Thus detection times are qualitatively different from likelihood ratings. It is important to note here that these response times had a clear category specific component [5]. To investigate this further, we trained models to predict detection response times, and generated their predictions on novel scenes from ADE20K[4]. Augmenting CNNs with these predictions barely improved performance (accuracy improvement: 0.34% for car, 0.87% for person), in contrast to the ~3% increase observed using likelihood predictions. We speculate that these gains are only incremental because detection times are strongly determined by target features [5] and only weakly by priors, and that target features are already captured reasonably well by CNNs.

Second, can the same performance benefits be obtained by augmenting CNNs with other models trained on target features or even target present scenes? To investigate this issue, we augmented CNNs with predictions of HOG-based models trained for car/person classification using a standard set of target-present and target-absent scenes. This yielded only a slight improvement in top-1 performance (0.4% for car & 0.1% for person) compared to the ~3% increase observed with human-derived priors.

Third, can similar performance benefits be obtained if CNNs are trained separately on target and background information? Recent studies suggest the answer to be in the affirmative. Specifically, modeling scene context and targets separately improves object detection above and beyond models trained on full scenes with objects embedded in context [32]. As expected, sampling local neighbourhoods of target objects instances has been shown to improve detection of objects with small real world sizes [33]. Even auxiliary tasks such as person action recognition [34], object segmentation [35] and predicting missing or wrongly located objects [36], can benefit when background regions are sampled independently. These studies complement our observation that augmenting object CNNs with human-derived context models can improve performance. We also speculate that models representing object and contextual information separately may also be more immune to overfitting to target features as is known to happen with very deep convolutional networks [3].

We surmise that there are more effective ways of integrating such human priors into deep convolutional architectures. Some promising avenues are attentional modules [2] and incorporating scale priors using skip layers [37]. It is possible that attentional mechanisms in humans are also optimized to yield benefits in object detection, since this is a core function of the human visual system. All these are interesting avenues for future study.

**METHODS**

*Participants.* Eleven subjects (3 female, 20-30 years old) participated in the task. All subjects had normal or corrected-to-normal vision and gave written informed consent to an experimental protocol approved by the Institutional Human Ethics Committee of the Indian Institute of Science, Bangalore.

*Stimuli.* For human behavioural experiments, we selected a total of 650 full colour real-world scenes with a resolution of 640 x 480 pixels (spanning 13.5° by 10.1° visual angle) containing neither cars nor people. Scenes included a wide range of natural and urban environments spanning many common scene categories (airport terminal, beach, botanical garden, bridge, coast, forest road, orchard, bamboo forest, bus station, cottage garden, driveway, forest, forest path, highway, hill, mountain, mountain path, mountain road, park, parking lot, picnic area, playground, rainforest, residential neighbourhood, river, runway, shipyard, ski lodge, ski resort, stage, taxiway, train station, tundra, valley, vegetable garden, village, waterfall, wheat field, woodland, workroom, parade ground). These scene categories are also well represented in the ADE20K dataset [4] which we have used for subsequent computational experiments. These 650 scenes also contained a variety of non-target objects. The number of times these objects occurred in these 650 scenes were: window (332), tree (327), pole (267), door (160), fence (149), sign (147), roof (147), text (103), lamppost (90), glass (82), cable (80), stripe (58), box (56), bush (47), stair (45), bench (42), rock (41), dustbin (36), flower-pot (35), lamp (29), flower (26), chair (26), entrance (23), cycle (22), table (20), boat (19), statue (17), hydrant (8), flag (8), wheel (7), animal (7), cone (6), bird (6), manhole-cover (5), cloud (5), bag (2).

*Procedure.* Subjects used a custom GUI interface created in Matlab®. They were instructed to assess how likely they thought a target could occur in the real scene if it was observed for a long time. They had to indicate this using a slider bar on the screen (with the two ends marked "very likely" to "very unlikely"). For every scene rated with non-zero likelihood for a category, the subject was asked to place a rectangular box to mark the most likely location and size at which the target would occur in the scene. For each scene, subjects had to indicate this for two target categories: cars and people in any order. The likelihood ratings were converted into a probability score by scaling them into the interval [0 1].

*Computational modelling of human expectations.* To understand the features that underlie human expectations, we extracted distinct types of visual information from each scene: targets, nontargets and scene context. Our approach is described and validated in detail elsewhere [5] and is summarized briefly below.

*Target features.* We extracted a total of 61 features from each scene. These features are templates of the visual appearance of cars and people across typical views and have been learned using an independent set of close cropped car and person images. We employed six models (2 categories x 3 views) based on Histograms of Oriented Gradients (HOG), which have been used previously to detect cars and people [38]. On convolution of the learned template with a scale pyramid of the scene, strong matches result in hits. We first thresholded the degree of match between the car/person template and a scene region at two levels, one is a tight threshold of -0.7 that has very few false alarms across the entire dataset and a second weaker threshold of -1.2 is set to allow for correct detections as well as false alarms. A diverse set of 31 attributes was extracted separately, once for car and once for person. These included the number of hits (n=1 feature) at high detector confidence $s(>-0.7)$, estimate of false-alarms (n=1 feature) computed as the difference between number of detections at strong ($>0.7$), average scale (area) of detected box (n=1 feature), and weak partial matches ($>-1.2$). Part-deformation statistics (n=16 features) were calculated by first normalising each detection to a unit square and finding the displacement of each detected part from the mean location of the part across all scenes in ours dataset. We also included eccentricity (n=5 levels from center of scene) and frequency of detected model types (n = 6, 2 categories x 3 views). Finally, an average detection score (n=1) was extracted from HOG detections in a scene. Feature vectors for car and person were then concatenated and used as the target feature vector (n=62). We found this summary of target features to be more informative than HOG histograms [22] computed on the same detected locations.

*Nontarget features.* These comprised binary labels corresponding to the presence/absence of the full set of objects that occurred across the set of 650 scenes. We avoided extracting image features from these objects since these could potentially be shared with target features. We explored the possibility of testing automated object detection using deep neural networks [21,39], but this yielded too many erroneous labels that would compromise model predictions. Example nontarget labels are shown in Figures 1-2. Some representative nontarget labels and their frequency in the dataset is, window (332), tree (327), pole (267), door (160), fence (149), sign (147), roof (147), text (103), lamppost (90), glass (82), cable (80), stripe (58), box (56), bush (47), stair (45), bench (42), rock (41), dustbin (36), flower-pot (35), lamp (29), flower (26), chair (26), entrance (23), cycle (22), table (20), boat (19), statue (17), hydrant (8), flag (8), wheel (7), animal (7), cone (6), bird (6), manhole-cover (5), cloud (5), bag (2).

*Coarse scene features.* These consisted of a combination of features encoded by the fc7 layer of a state-of-art deep convolutional network (CNN) optimized for scene categorisation [30] together with the coarse spatial envelope GIST operator [18]. We included GIST features because they improved model predictions for horizontal locations of objects and marginally improved overall performance. In both cases, features were extracted by giving as input to each model a blurred version of the scene. The blurred scene was obtained by convolving the original scene with a low pass Gaussian filter ($\sigma = 20$ pixels), such that objects and their parts were no longer recognizable. To confirm that target or nontarget information was no longer present in these images, we took blurred scenes with and without cars/people and asked whether object-based detectors [38] could correctly identify the scenes containing targets. This yielded poor detection accuracy (average accuracy: <5% for both car and person detectors across 100 randomly chosen scenes).

*Model fitting and performance evaluation.* We sought to assess whether human likelihood judgments on scenes could be predicted using target, nontarget and coarse scene features or a combination of these channels. To this end we fit models based on every possible subset of these channels. To identify the best model, we selected the model that outperformed all other models in terms of the match between observed likelihood ratings and cross-validated model predictions. We equated the complexity of each feature channel by projecting each subset of features along their first 20 principal components. This typically captured over 85% of variance across 650 scenes for each of the three information channels and provided a compact description of the features in each channel.

All models were fit with linear regression of the form $\mathbf{y} = \mathbf{Xb}$, where $\mathbf{y}$ is the vector of likelihood ratings (likelihood/x-location/y-location/scale/aspect-ratio), $\mathbf{X}$ is a matrix whose rows contain features for each scene derived from targets, nontargets and coarse scene structure and $\mathbf{b}$ is a vector of unknown weights representing the contribution of each column in $\mathbf{X}$. We used standard linear regression to solve this equation. We tested all models for their ability to predict average ratings on novel scenes using 5-fold cross-validation. All models were trained and tested on scenes that were devoid of cars as well as people and hence only predict the human beliefs about car or person attributes such as likelihood of presence, location or scale. We concatenated model predictions on the cross-validation test sets and calculated the correlation with the observed ratings obtained from the behavioural experiment. A perfect agreement between predicted and observed ratings would yield a correlation coefficient of 1 with a high statistical significance (i.e. $p < 0.05$ of observing this correlation by chance). In contrast, non-informative model predictions would result in near-zero correlations that are typically not statistically significant.

*Noise ceiling estimates.* To estimate an upper bound for model performance, we reasoned that model performance cannot exceed the reliability of the data. We estimated this reliability by calculating the correlation coefficient between average per-scene ratings between two randomly chosen groups of subjects, and applying a correction to account for the fact that this correlation is obtained between two halves of the data rather than on the full dataset. This correction, known as the Spearman-Brown correction, is given by *rc = 2r/(r+1)*, where *r* is the split-half correlation.

*Augmenting CNNs with human-derived expectations.* We selected two state-of-the-art CNNs for testing. The first CNN was similar to the BVLC reference classifier [3] that has a mean average precision (mAP) of 72% on the PASCAL VOC 2007 dataset [31]. Hereafter we refer to this CNN as Alexnet. The second CNN has an inbuilt attention module and generates region proposals on which detection is carried out [2]: this model has 73.2% mAP on the same dataset [31]. We gave the highest possible benefit to this model by selecting the most confidently detected instance within every scene and for each category. Hereafter we refer to this CNN as RCNN. To evaluate object detection performance, we used images from the recently released ADE20K [4] scene dataset [4]. This dataset contains over 20,000 real-world scenes with 5601 scenes containing people and 3245 scenes containing cars. The chosen scenes have high variability in composition of scenes as well as visual attributes of targets. For negative examples, we randomly sampled matching sets of car absent (n = 3245) and person absent scenes (n=5601). We also selected a restricted subset of 372 scenes from the 3470 scenes containing cars, by matching scene types present in our reference set of 650 car-person absent scenes (see Methods). Likewise, we also selected a subset of 306 scenes from the larger set of 5601 scenes containing people. We have further summarized these selection choices in the Supplementary Table 2.

**ACKNOWLEDGMENTS**

This work was funded through the ITPAR collaborative grant (to S.P.A. & M.V.P.) from the Department of Science and Technology, Government of India and the Province of Trento. HK was supported by a postdoctoral fellowship from the DST Cognitive Science Research Initiative, Government of India. MVP was supported by funding from the European Research

Council (ERC) under the European Union's Horizon 2020 research and innovation programme (grant agreement No 725970). SPA was supported by Intermediate and Senior Fellowships from the Wellcome Trust - DBT India Alliance.

## AUTHOR CONTRIBUTIONS
HK, MVP & SPA designed experiments; HK collected and analysed data; HK & SPA wrote manuscript with inputs from MVP. The authors declare no competing financial interests

**FIGURE LEGENDS**

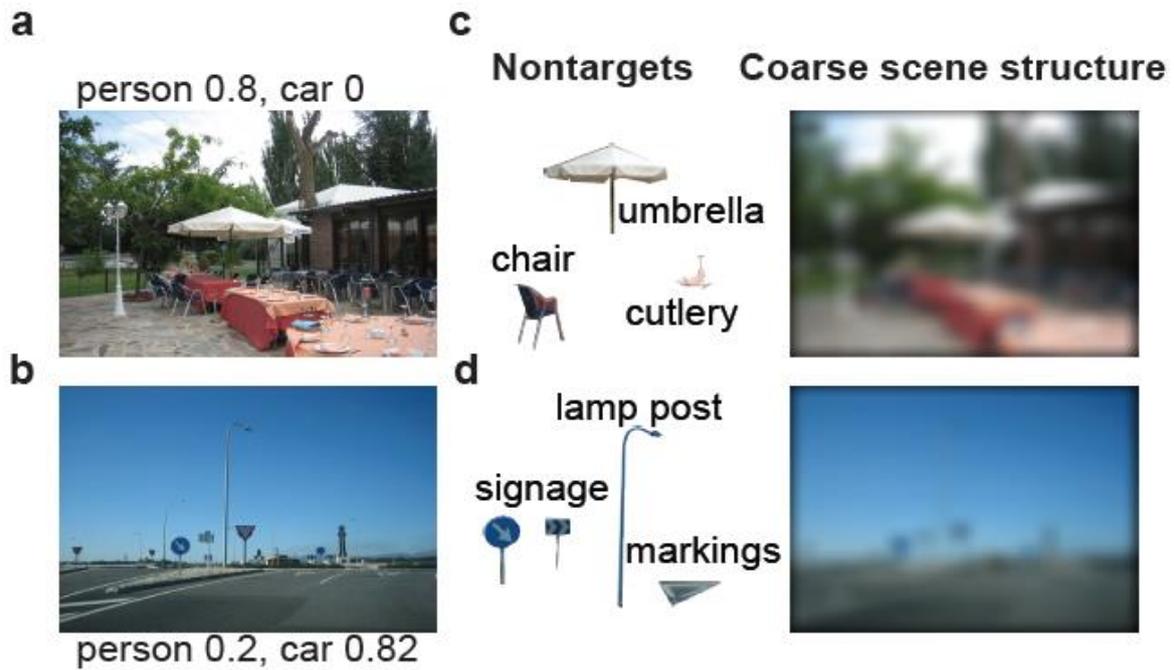

**Figure 1.** (a) Example scene rated by subjects as likely to contain people but not cars. (b) Example scene with high car and low person likelihood (c-d) show nontarget objects. We modeled these expectations using person/car features (not shown), nontarget objects (*middle*) and coarse scene structure (*right*).

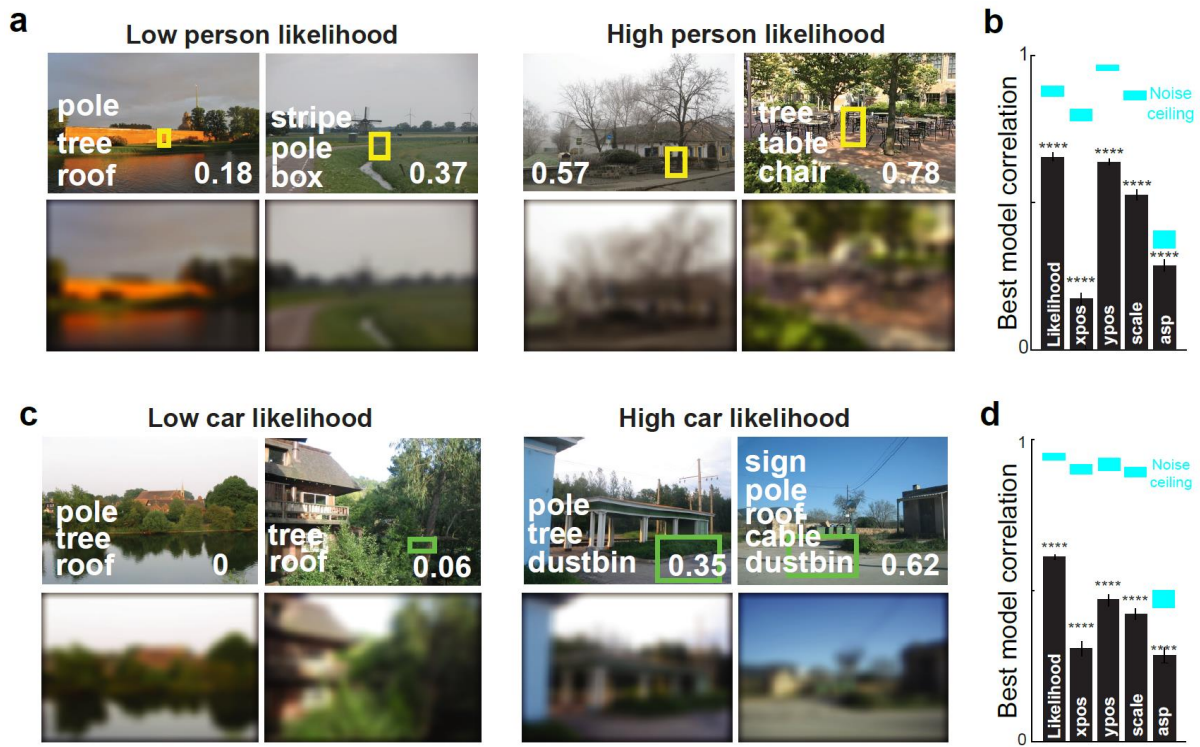

**Figure 2.** (a) Example scenes rated by subjects as having low and high person likelihood (top row) with nontarget labels and coarse scene structure (*bottom row*). Yellow boxes indicate the average location and scale at which a person was marked as most likely to occur in each scene by subjects (b) Correlation between best model (NC: nontargets and coarse scene features) predictions for likelihood, and the most likely horizontal position (xpos), vertical position (ypos), scale and aspect-ratio (asp) at which a person might occur in the scene. Cyan regions above each bar represent the reliability of the human data (mean± std of corrected split-half correlation; see text) (c-d) Analogous plots for car likelihood data.

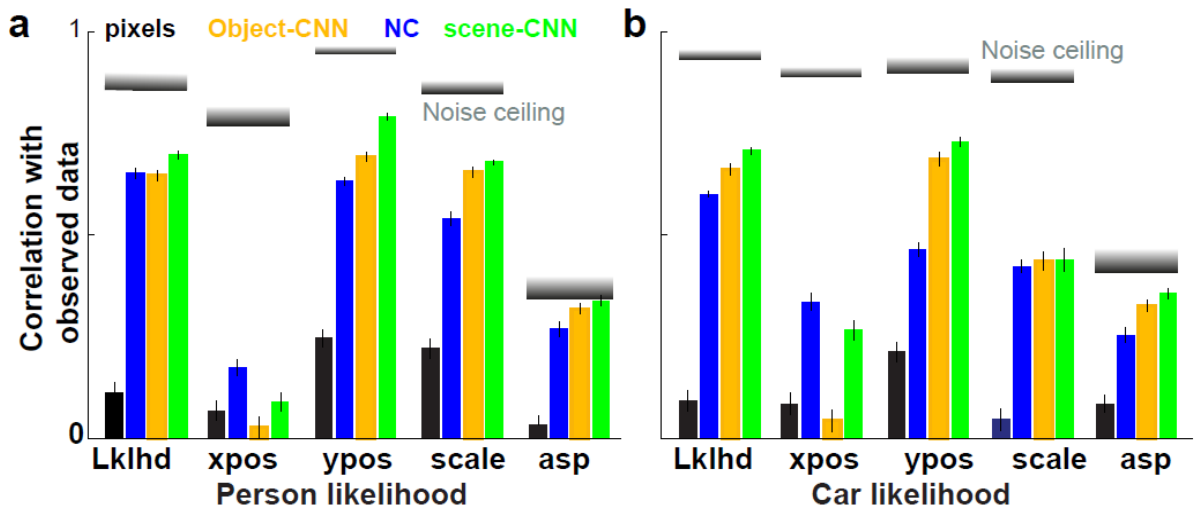

**Figure 3.** Comparison with other models. (a) Model performance on person likelihood data for raw pixels, nontarget+coarse scene features, object-CNN and scene-CNN. The object-CNN was pre-trained for 1000-way object classification and the scene-CNN was pre-trained for 205-way scene classification. Shaded gray bars represent the noise ceiling for each type of data (mean ± std). (b) Model performance for car likelihood data. *Lklhd*: Likelihood.

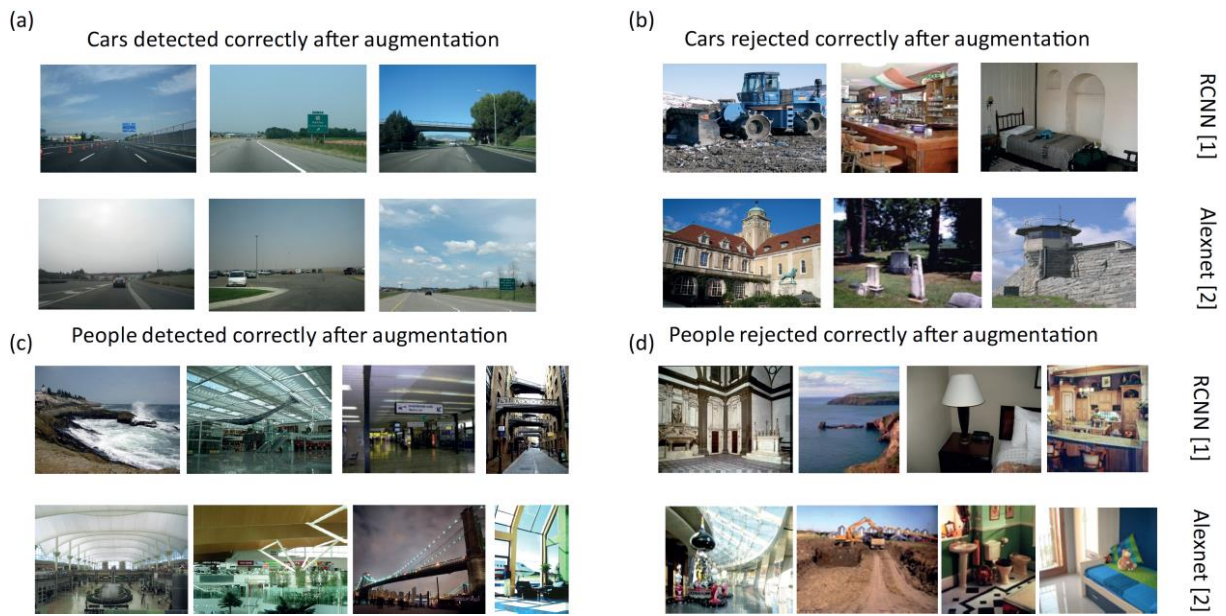

**Figure 4.** Augmenting CNNs with human expectations helps to accept low confidence detections (left) and reject false alarms (right). (a) Scenes containing small and hard to detect cars, these scenes are correctly classified as containing cars after augmentation with human derived priors (b) car false alarms that are correctly rejected after augmentation with human

derived priors. (c) Scenes with multiple people at small scales and unusual locations (d) scenes devoid of people but falsely classified as person present by CNNs. All pictures best seen in high resolution in the digital version.

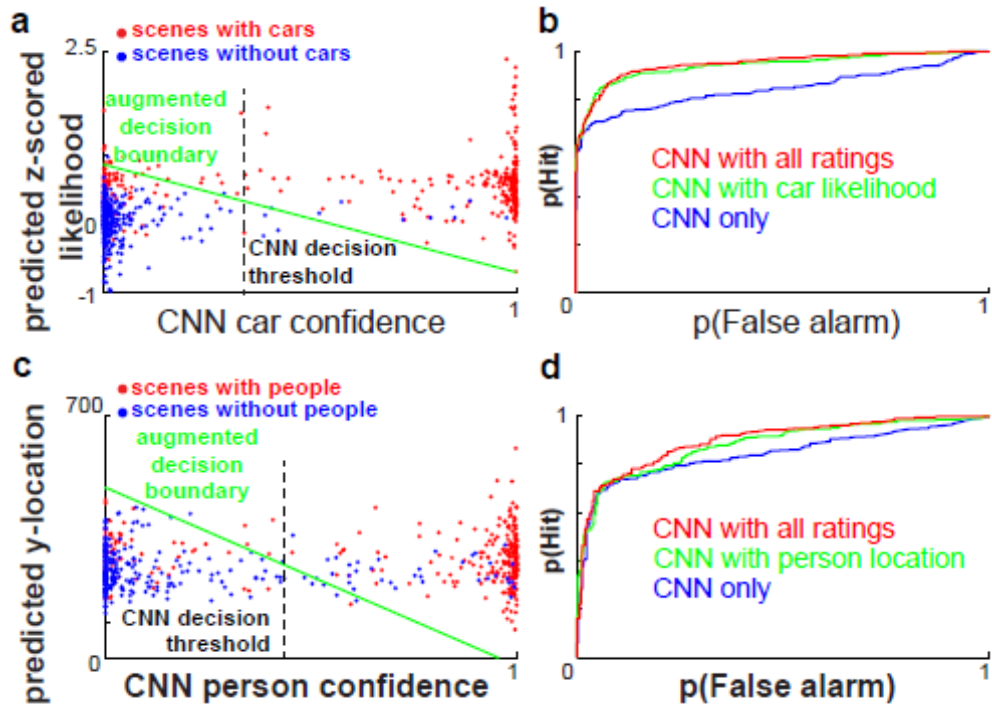

**Figure 5.** Augmenting CNNs with human expectations improves performance. (a) Classifier boundaries before (dashed line) and after (green) augmenting a CNN with predicted car likelihood ratings (b) ROC curves for the CNN, CNN with car likelihood & CNN with all car ratings. (c-d) Analogous plots for person detection augmented by predicted vertical location.

# Supplementary Tables

| Model | Person data | | | | Car data | | | |
|---|---|---|---|---|---|---|---|---|
| | xpos | ypos | scale | asp | xpos | ypos | scale | asp |
| *Ceil* | *0.79±0.02* | *0.96±0.01* | *0.86±0.01* | *0.36±0.03* | *0.9±0.01* | *0.91±0.2* | *0.89±0.02* | *0.47±0.03* |
| **TNC** | 0.17±0.02# | 0.63±0.01# | 0.52±0.02* | 0.28±0.02# | 0.30±0.02 | 0.45±0.02# | 0.41±0.02# | 0.27±0.02# |
| **T** | 0.00±0.02* | 0.10±0.02* | 0.09±0.02* | 0.09±0.02* | 0.10±0.03* | 0.08±0.02* | 0.08±0.03* | 0.14±0.02* |
| **N** | 0.10±0.02* | 0.40±0.01* | 0.44±0.01* | **0.29±0.01** | 0.03±0.03* | 0.21±0.02* | 0.22±0.02* | 0.13±0.02* |
| **C** | **0.18±0.02** | 0.58±0.01* | 0.46±0.01* | 0.19±0.02* | **0.33±0.02** | 0.43±0.01 | **0.42±0.02** | **0.30±0.02** |
| **TN** | 0.09±0.02* | 0.40±0.01* | 0.42±0.02* | 0.27±0.02# | 0.00±0.02* | 0.24±0.02* | 0.24±0.02* | 0.17±0.02* |
| **TC** | 0.17±0.02# | 0.59±0.01* | 0.46±0.01* | 0.20±0.02* | 0.28±0.02* | 0.42±0.02 | 0.40±0.02# | 0.28±0.02# |
| **NC** | 0.17±0.02# | **0.63±0.01** | **0.54±0.02** | 0.29±0.02# | 0.33±0.02# | **0.46±0.02** | 0.42±0.02# | 0.30±0.02# |

**Table S1. Model performance on predicting likely location, scale and aspect ratio.** Models based on various combinations of features were trained separately to predict the horizontal position (xpos), vertical position (ypos), area (scale) and aspect ratio (asp) at which a person was most likely to occur in the scene. The best model in each case is indicated using **bold face**. All other conventions are as in Table 1.

| Scene subsets from ADE20K[3] | Attributes | | | |
|---|---|---|---|---|
| | #Scenes | #Car present | #People present | Scene categories |
| **Car (C)** | 6940 | 3470 | 0 | All scene categories in ADE20K [3] |
| **Car matched (CM)** | 744 | 372 | 0 | Matched to 650 scenes in behavioral experiment |
| **Person (P)** | 11202 | 0 | 5601 | All scene categories in ADE20K [3] |
| **Person matched (PM)** | 612 | 0 | 306 | Matched to 650 scenes in behavioral experiment |

**Table S2. Scenes used to evaluate CNN augmentation with human derived priors.**